\documentclass{article}
\usepackage{spconf,amsmath,graphicx}

\usepackage{times}

\usepackage{graphicx} 
\usepackage{subfig} 

\usepackage{algorithm}
\usepackage{algorithmic}

\usepackage{hyperref}

\usepackage{multirow}

\def \puntaco {\bullet}

\usepackage[utf8]{inputenc} 
\usepackage[T1]{fontenc}    
\usepackage{url}            
\usepackage{booktabs}       
\usepackage{amsfonts}       
\usepackage{microtype}      

\usepackage{pgfplots}
  \pgfplotsset{compat=newest}
  \pgfplotsset{major grid style={dashed,line width=0.5pt}}
  \pgfplotsset{/pgf/number format/1000 sep={}}
  \usetikzlibrary{plotmarks}
  \usepgfplotslibrary{patchplots}
  \usepackage{grffile}
  \usepackage{tikz}
\usetikzlibrary{bayesnet}
\pgfplotsset{major grid style={dashed,line width=0.5pt}}
\pgfplotsset{/pgf/number format/1000 sep={}}

\usepackage{lmodern}

\usepackage{adjustbox}

\usepackage{amsmath}
\usepackage{color}

\def \X {\mathbf{X}}
\def \Z {\mathbf{Z}}
\def \A {\mathbf{B}}

\usepackage{amssymb}
\usepackage{amsgen,amsbsy,amsopn}
\usepackage{bm}

\usepackage{bbm}
\usepackage{epstopdf}

\usepackage{array}
\newcolumntype{L}[1]{>{\raggedright\let\newline\\\arraybackslash\hspace{0pt}}m{#1}}
\newcolumntype{C}[1]{>{\centering\let\newline\\\arraybackslash\hspace{0pt}}m{#1}}
\newcolumntype{R}[1]{>{\raggedleft\let\newline\\\arraybackslash\hspace{0pt}}m{#1}}

\title{Sparse Three-parameter Restricted Indian buffet process \\ for Understanding International Trade}
\name{Melanie F. Pradier$^1$, Viktor Stojkoski$^2$, Zoran Utkovski$^2$, Ljupco Kocarev$^2$, and Fernando Perez-Cruz$^1$\thanks{Thanks to the European Union 7th Framework Programme through the Marie Curie ITN ``Machine Learning for Personalized Medicine'' MLPM2012 for funding.}}
\address{$^1$ University Carlos III in Madrid; $^2$ Macedonian Academy of Sciences and Arts}
\begin{document}
%
\maketitle
\begin{abstract}
This paper presents a Bayesian nonparametric latent feature model specially suitable for exploratory analysis of high-dimensional count data. We perform a non-negative doubly sparse matrix factorization that has two main advantages: not only we are able to better approximate the row input distributions, but the inferred topics are also easier to interpret. By combining the three-parameter and restricted Indian buffet processes into a single prior, we increase the model flexibility, allowing for a full spectrum of sparse solutions in the latent space. We demonstrate the usefulness of our approach in the analysis of countries' economic structure. Compared to other approaches, empirical results show our model's ability to give easy-to-interpret information and better capture the underlying sparsity structure of data. 
\end{abstract}

\vspace{-0.2cm}
\section{Introduction}

Exploration in high-dimensional data needs to balance predictive accuracy with interpretability \cite{jordan_message_2011}. When collaborating with experts in other fields, the primary goal is often not only reducing some error measure, but rather understanding the structure of data \cite{doshi-velez_towards_2017, kim_mind_2015, vellido_making_2012}.
 The data exploration phase can then be turned into a data exploitation phase via policy recommendations, medical protocols or as a further improved discriminative model~\cite{ribeiro_why_2016}. 

Data exploration comes in different forms. PCA and factor analysis are linear methods that provide non-sparse solutions with strong Gaussianity assumptions. Local linear embedding \cite{roweis_nonlinear_2000}, isomap \cite{tenenbaum_global_2000} and Gaussian process latent variable models \cite{lawrence_probabilistic_2005} learn non-linear manifolds in high dimensional spaces with non-sparse features; non-negative matrix factorization \cite{hoyer_non-negative_2004} provides a low dimensional sparse representation of the data. Also, Bayesian nonparametric (BNP) models  can be used for clustering \cite{archambeau2015latent} and sparse feature analysis \cite{griffiths_indian_2011}, in which the underlying latent dimension is unknown. Many applications have benefited from sparse data exploration models, like 
computer vision~\cite{wright_sparse_2010}, 
genomics~\cite{knowles_nonparametric_2011,chen_phylogenetic_2013}, 
psychiatry ~\cite{ruiz_bayesian_2014}, or sports~\cite{pradier_prior_2016}.

\begin{figure}[]
	\centering
	\hspace{-0.6cm}\input{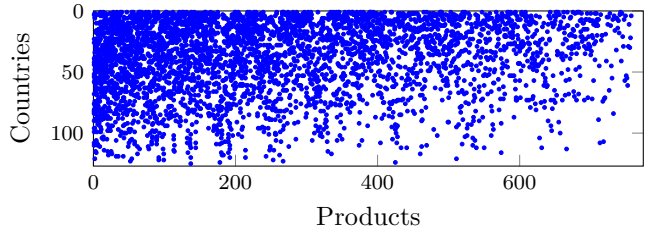}
	\vspace{-0.3cm}\begin{footnotesize}\caption{\textbf{Data considered in this paper.} A non-zero entry reflects a relative advantage of a country at exporting a given product. Note the triangular structure.
	} \label{fig:1}
    \end{footnotesize}
    \vspace{-0.5cm}
\end{figure}

This paper presents a non-negative matrix factorization model specially suitable for high-dimensional count data. Our model provides easy-to-interpret features and captures sparsity structure in data.
This is illustrated in the context of international trade and economic growth
~\cite{hidalgo_product_2007, hausmann_atlas_2014,stojkoski_impact_2016}.
%
Data considered in this paper is approximately triangular after reordering of rows and columns, as shown in Fig.~\ref{fig:1}: here, countries have different diversity degrees in their export portfolios, and thus different trade strategies and skills. Our objective is then to capture such triangular structure, e.g., discover the underlying capabilities of countries, and their relationships. 
%
%
%

We rely on two different extensions of the Indian buffet process (IBP) to learn a potentially infinite number of latent features. 
The three-parameter IBP allows for different degrees of sharing between features~\cite{teh_indian_2009}, whereas the restricted IBP allows for a general distribution over the number of active features per row~\cite{doshi-velez_restricted_2015}. We combine both elements into a sparse Poisson factorization scheme. In the trade context, our model is able to represent different kinds of realities, from a world in which countries with few skills focus on different types of products, to a world in which poor countries have a strong overlap in export skills. We also allow for data points to exhibit different number of active features, e.g., we expect poor countries to have very little features active, while developed countries might have almost all features active. Our contributions include a novel BNP model for count data, a corresponding inference algorithm, and extensive empirical validation  in the context of trade data.

\vspace{-0.2cm}
\section{Theoretical Background}\label{sec:background}

\subsection{Indian buffet process (IBP)}\label{sec:IBP}

The IBP is a stochastic process defining a probability distribution over equivalence classes of sparse binary matrices with a finite number of rows and unbounded number of columns~\cite{griffiths_indian_2011}. Although the number of columns is potentially infinite, only a finite number of those will contain non-zero entries due to the finite nature of the observed data. 
 The IBP can be derived taking the limit as $K \rightarrow \infty$ of a finite binary matrix $\Z \in \{0,1\}^{N \times K}$, where $N$ is the number of observations, and $K$ is the number of latent features. Each element $z_{nk}$ is generated as:
\vspace{-0.1cm}
\begin{eqnarray}
\pi_k &\sim & \text{Beta}(\alpha/K, 1), \nonumber \\
z_{nk} &\sim & \text{Bernoulli}(\pi_k)
\end{eqnarray}
where  $\pi_k$ is the probability of observing a non-zero value in column~$k$, $\Z_{n \puntaco}$~is the $n$-th row for sample~$n$ and $\Z_{\puntaco k}$ is the $k$-th column for feature~$k$. We say that a feature $k$ is active for sample $n$ if $z_{nk}=1$. When $K \rightarrow \infty$, the above finite model tends to the IBP, denoted by:
$\mathbf{Z} \sim \text{IBP}(\alpha)$,
 where $\alpha$ is the mass parameter controlling the a priori activation probability of new features. 

In the IBP, the expected number of active features per row is distributed according to $\mathrm{Poisson}(\alpha)$ and the total number of active features $K^{+}$, i.e., number of columns with non-zero entries,
is distributed as $\mathrm{Poisson}\big(\alpha\sum_{i=1}^N\left(\frac{1}{i}\right)\big)$.
%
%
The single scalar parameter $\alpha$ has thus an effect on both the density (total number of ones) and sparsity structure (position of the non-zero values within $\Z$).
Such assumption might be too restrictive in general data exploration tasks, failing to capture situations such as a high number of latent features with low activation levels, or varying degrees of per-row sparsity in the latent matrix.

\subsection{Extensions of the IBP}\label{sec:extensions}

\paragraph*{Three-Parameter IBP (3P-IBP)}
As its name indicates, the 3P-IBP can be fully specified by three parameters~\cite{teh_indian_2009}: $\alpha$ is the same mass parameter from the IBP, $\sigma \in [0,1)$ controls the power-law behavior of the model (weight decay), and $c>-\sigma$ is the concentration parameter that affects the {\it a priori} number of ones per column (sharing degree across features). When $c=1$ and $\sigma=0$, we recover the standard IBP model.

%
%
By introducing parameters $c$ and $\sigma$, the latent matrix $\Z$ (defined in~\ref{sec:IBP}) has a more flexible sparsity structure, regardless of the sparsity density which is controlled by $\alpha$.  The 3P-IBP gives more flexibility on the feature weights, but has the disadvantage that the number of ones per-row is still Poisson distributed a priori for all data points, which might not be desirable in all scenarios. This problem can be directly addressed by the R-IBP.

\paragraph*{Restricted IBP (R-IBP)}
%
%
The recently developed R-IBP allows for an arbitrary prior distribution $f$ over the number of active features per row~\cite{doshi-velez_restricted_2015}.
%
The R-IBP has two degrees of freedom $\alpha$ and $f$ to respectively control for sparsity degree and sparsity structure of $\Z$. 
%
%
 The intuition behind the R-IBP is easy to explain in the commonly used culinary metaphor for IBPs, where rows design customers, and columns refer to dishes in an Indian buffet~\cite{griffiths_indian_2011}. Customers in the R-IBP have varying degrees of hunger: some of them sample from many dishes in the buffet (the non-zero values), while others only taste a reduced set of dishes. This is generally convenient to model structured data, e.g., international trade, where developed countries are known to have more assets, and thus are expected to exhibit a higher number of latent features (capabilities) compared to poor countries.

\section{Our Approach}\label{sec:S3R-IBP}

\paragraph*{Modeling}
Let $\X \in \mathbb{N}^{N \times D}$ be our input matrix of $N$ data points and $D$ dimensions.
  We build an infinite latent feature model for count data with Poisson likelihood and Gamma-distributed factors:
\vspace{-0.2cm}
\begin{eqnarray}\label{eq:lik_SPFM}
x_{nd} & \sim & \mathrm{Poisson}\big(\Z_{n \puntaco} \A_{\puntaco d} \big),\\
B_{kd} & \sim & \mathrm{Gamma}\big( \alpha_B , \frac{\mu_B}{\alpha_B}),
\end{eqnarray}
where $\Z$ is a binary matrix, and $\alpha_B$ and $\mu_B$ are the shape and mean parameters of the prior Gamma distribution for each element of matrix $\A$. Sparsity in matrix $\A$  can be induced simply by choosing $\alpha_B \ll 1$. Both $\Z$ and $\A$ are then non-negative and sparse, which makes the inferred latent variables particularly interpretable.
 To decouple sparsity density and sparsity structure in $\Z$, we combine the advantages of both the R-IBP and 3P-IBP into a single prior,
\vspace{-0.1cm}
\begin{eqnarray}\label{eq:lik_SPFM2}
\Z & \sim & \text{3R-IBP}(\alpha, c,\sigma,f),\label{eq:RIBP}
\end{eqnarray}
where $\alpha, c, \sigma$, and $f$ refer to the parameters defined in Sec.~\ref{sec:extensions}. We rely on a negative binomial distribution for $f$, which is best understood as an overdispersed Poisson. Hence it will naturally allow for countries to exhibit a much variable range of active features. 
%
%
%
%
We refer to the model as \textit{Sparse Three-parameter Restricted Indian buffet process} (S3R-IBP). It can be seen as a probabilistic extension of non-negative matrix factorization where the number of latent features is not fixed a priori, both matrices are sparse, and soft-constraints on the latent sparsity structure are imposed through a more flexible prior.

In the trade context, we have $N$ countries, $D$ products and $K^{+}$ non-empty latent features to be inferred. A given row $\mathbf{Z}_{n \puntaco}$ captures which latent features (skills) are active for country~$n$. Matrix $\A$ represents the effect of each latent feature on every product. For instance, if a latent feature $k$ is active for a certain country, all products having high values in vector $\A_{k \puntaco}$ will be more likely to be exported by that country.

\vspace{-0.2cm}
\paragraph*{Inference}\label{sec:inference}
Since exact computation of the posterior distribution for the latent variables is intractable, we resort to a Markov Chain Monte Carlo (MCMC) approach.
Our algorithm uses Gibbs sampling together with Metropolis-Hasting (MH). 
Following~\cite{gopalan_scalable_2013}, we introduce the auxiliary variables $x^{'}_{nd,1}, \ldots, x^{'}_{nd,K}$ for each observation $x_{nd}$ such that $x_{nd} = \sum^K_{k=1}x^{'}_{nd,k}$, and $x^{'}_{nd,k} \sim \text{Poisson}(Z_{nk}B_{kd})$ for $k=1,\ldots,K$.
%
%
Given such auxiliary variables, the model is conditionally conjugate, and a Gibbs sampler can be derived straightforwardly.
The complete sampling algorithm is described in Alg.~\ref{alg:sampling}. 
 
\section{Experiments}

%
Two publicly available trade datasets, the SITC and HS, are considered for the year 2010. The data represents the Revealed Comparative Advantage (RCA) of countries, a normalized common measure in economics~\cite{balassa_purchasing-power_1964} already illustrated in Fig.~\ref{fig:1}. 
 Simulations are run for 10 different train-test splits with a proportion of 90-10\% entries. The MCMC burn-in period is 30,000 iterations, and results are averaged using the last 1,000 posterior samples.


%

\begin{algorithm}[t]
\caption{A single iteration of the MCMC inference procedure for the S3R-IBP model.}
\label{alg:sampling}
\begin{algorithmic}[1]
 \STATE Sample each element of $\mathbf{Z}$ using inclusion probabilities~\cite{aires_algorithms_1999,doshi-velez_restricted_2015}.
 \STATE Sample latent measure $\boldsymbol{\pi}$ using MH steps~\cite{doshi-velez_restricted_2015}. 
 \STATE Sample each element of $\A$ and $\X^{'}$ from their conditional distributions (conjugate priors).
 \STATE Sample hyperparameter $\alpha$ according to~\cite{archambeau2015latent}.
\end{algorithmic}
\end{algorithm}
 \vspace{-0.1cm}

\vspace{-0.2cm}
\paragraph*{Model hyperparameters} We choose $f = \text{Negative-}$ $\text{Binomial}(r,p)$, with $r=[1,2]$, and $p=[0.1, 0.3, 0.5]$. Results were equivalent using any of these priors. Here, we report results for $r=1$ and $p=0.1$. We also ran experiments for each combination of $c=[1, 10, 20, 50]$ and $\sigma=[0,0.25, 0.5, 0.75, 1]$. Parameter $c$ was found to be more influential than $\sigma$. 
We here report the best setting $c=50$, and $\sigma = 1$. Hyperparameters for the Gamma prior over $\alpha$ are shape and scale equal to one. Finally, $\alpha_B$ is set to 0.01 to induce sparsity, and $\mu_B=1$.
%

\subsection{Quantitative evaluation}

Table~\ref{tab:performance} compares our model against probabilistic matrix factorization (MF) \cite{Salakhutdinov08}, non-negative MF (NMF) \cite{Schmidt09}, IBP~\cite{griffiths_indian_2011}, and sparse IBP (S-IBP) which uses $\alpha_B < 1$, in terms of predictive accuracy and interpretability strength. 

%
%

\begin{table*}
\footnotesize
\begin{center}
\subfloat[2010 SITC database ($N=126$, $D=744$, 16k non-zero values, 17\% sparsity)]{
{\begin{tabular}{@{}rccccc@{}}\toprule Metric & MF & NMF & IBP & S-IBP & S3R-IBP \\\toprule\midrule

Log Perplexity & $1.68 \pm 0.01$ & $1.61 \pm 0.01$ & $\mathbf{1.59 \pm 0.04}$ & $3.26 \pm 0.17$ & $1.62 \pm 0.01$\\ 
Coherence & $-264.60 \pm 4.74$ & $-263.27 \pm 7.45$ & $-149.36 \pm 7.56$ & $-178.44 \pm 4.50$ & $\mathbf{-140.51 \pm 2.73}$ \\ 
\toprule \end{tabular}}{}


}
\vspace{-0.2cm}
\subfloat[2010 HS database ($N=123$, $D=4890$, 77k non-zero values, 13\% sparsity)]
{\begin{tabular}{@{}rccccc@{}}\toprule Metric & MF & NMF & IBP & S-IBP & S3R-IBP \\\toprule\midrule

Log Perplexity & $1.48 \pm 0.01$ & $\mathbf{1.47 \pm 0.01}$ & $1.58 \pm 0.01$ & $2.56 \pm 0.12$ & $1.57 \pm 0.02$\\ 
Coherence & $-264.73 \pm 3.11$ & $-264.67 \pm 6.22$ & $-148.91 \pm 10.57$ & $-168.39 \pm 13.16$ & $\mathbf{-134.51 \pm 4.43}$\\ 
\toprule \end{tabular}}{}
\end{center}
\vspace{-0.6cm}
\caption{\textbf{Quantitative evaluation of accuracy and interpretability.} S3R-IBP beats MF, NMF, IBP, and S-IBP in terms of topic coherence while retaining similar predictive accuracy (in terms of test log-perplexities).}\label{tab:performance}
\end{table*}

\vspace{-0.1cm}
\paragraph*{Accuracy}
All models present similar perplexity (the lower, the better), except S-IBP, in which the sparseness restriction degrades its performance significantly. S3R-IBP has the same sparsity constraint, but its more flexible prior compensates the penalty in perplexity, leading to a performance close to the non-sparse models, i.e. MF and IBP. The S3R-IBP match the perplexity performance of non sparse methods, but keeping the results interpretable.


\vspace{-0.1cm}
\paragraph*{Interpretability}
To assess semantic quality, we rely on coherence~\cite{
 fan_promoting_2017}, which is an often-used metric in topic modeling literature. 
 The closer coherence is to zero, the better. 
S3R-IBP outperforms IBP and S-IBP by far, making it specially suitable for data exploration in high-dimensional count scenarios. The non-sparse methods present a very low coherence, as expected.

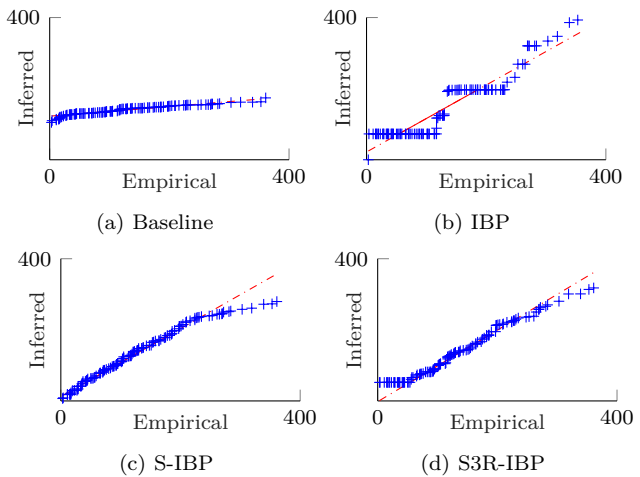
\begin{figure}[b!]
	\vspace{-0.6cm}
	\centering
	\hspace{-0.3cm}\subfloat[Baseline]{
%
%
\definecolor{mycolor1}{rgb}{0.00000,0.44700,0.74100}%
\begin{tikzpicture}

\begin{axis}[%
width=0.37\columnwidth,
height=0.22\columnwidth,
scale only axis,
xlabel style={at={(axis description cs:0.5,-0.03)},anchor=north},
every x tick label/.append style={font=\footnotesize},
xmin=0,
xmax=400,
xtick={  0, 400},
xlabel style={font=\footnotesize\color{white!15!black}},
xlabel={Empirical},
every y tick label/.append style={font=\footnotesize},
ymin=0,
ymax=400,
ytick={400},
ylabel style={font=\footnotesize\color{white!15!black}},
ylabel style={at={(axis description cs:-0.03,.5)},anchor=south},
ylabel={Inferred\tiny},
axis background/.style={fill=white},
axis x line*=bottom,
axis y line*=left
]
\addplot [color=red, dashdotted, forget plot]
  table[row sep=crcr]{%
3	123.045454545455\\
361	171.863636363636\\
};
\addplot [color=red, forget plot]
  table[row sep=crcr]{%
54	130\\
186	148\\
};
\addplot [color=mycolor1, draw=none, mark=+, mark options={solid, blue}, forget plot]
  table[row sep=crcr]{%
3	105\\
4	110\\
11	112\\
11	114\\
15	114\\
16	115\\
16	116\\
16	118\\
17	118\\
17	120\\
19	120\\
20	121\\
23	123\\
27	124\\
28	125\\
29	126\\
33	127\\
34	127\\
35	127\\
36	127\\
37	127\\
41	128\\
42	128\\
43	128\\
44	129\\
44	130\\
45	130\\
50	130\\
52	130\\
54	130\\
56	130\\
57	130\\
59	130\\
65	131\\
66	131\\
70	131\\
71	131\\
73	132\\
76	132\\
80	133\\
83	133\\
86	133\\
88	133\\
89	134\\
92	134\\
93	134\\
93	135\\
95	135\\
95	136\\
101	136\\
102	136\\
103	136\\
105	136\\
105	137\\
113	137\\
114	139\\
117	139\\
117	141\\
117	142\\
119	142\\
121	143\\
123	143\\
127	143\\
128	143\\
129	143\\
130	144\\
135	144\\
137	144\\
138	145\\
143	145\\
148	145\\
151	145\\
153	145\\
155	145\\
158	146\\
164	146\\
164	147\\
167	147\\
173	147\\
175	148\\
178	148\\
182	148\\
185	148\\
186	148\\
186	149\\
190	149\\
197	150\\
198	150\\
199	151\\
201	151\\
204	152\\
209	152\\
210	152\\
216	153\\
224	153\\
227	153\\
229	154\\
231	154\\
235	154\\
248	154\\
253	155\\
261	155\\
265	155\\
269	155\\
270	157\\
272	157\\
281	157\\
284	158\\
303	161\\
319	162\\
339	162\\
353	163\\
361	174\\
};
\end{axis}

\end{tikzpicture}%
}\subfloat[IBP]{
%
%
\definecolor{mycolor1}{rgb}{0.00000,0.44700,0.74100}%
\begin{tikzpicture}

\begin{axis}[%
width=0.37\columnwidth,
height=0.22\columnwidth,
scale only axis,
xlabel style={at={(axis description cs:0.5,-0.03)},anchor=north},
every x tick label/.append style={font=\footnotesize},
xmin=0,
xmax=400,
xtick={  0, 400},
xlabel style={font=\footnotesize\color{white!15!black}},
xlabel={Empirical},
every y tick label/.append style={font=\footnotesize},
ymin=0,
ymax=400,
ytick={400},
ylabel style={font=\footnotesize\color{white!15!black}},
ylabel style={at={(axis description cs:-0.03,.5)},anchor=south},
ylabel={Inferred\tiny},
axis background/.style={fill=white},
axis x line*=bottom,
axis y line*=left
]
\addplot [color=red, dashdotted, forget plot]
  table[row sep=crcr]{%
3	24.6450442703947\\
361	361.862501004738\\
};
\addplot [color=red, forget plot]
  table[row sep=crcr]{%
54	72.684402631991\\
186	197.02156545024\\
};
\addplot [color=mycolor1, draw=none, mark=+, mark options={solid, blue}, forget plot]
  table[row sep=crcr]{%
3	7.88531906437129e-10\\
4	72.684402631991\\
11	72.684402631991\\
15	72.684402631991\\
16	72.684402631991\\
17	72.684402631991\\
19	72.684402631991\\
20	72.684402631991\\
23	72.684402631991\\
27	72.684402631991\\
28	72.684402631991\\
29	72.684402631991\\
33	72.684402631991\\
34	72.684402631991\\
35	72.684402631991\\
36	72.684402631991\\
37	72.684402631991\\
41	72.684402631991\\
42	72.684402631991\\
43	72.684402631991\\
44	72.684402631991\\
45	72.684402631991\\
50	72.684402631991\\
52	72.684402631991\\
54	72.684402631991\\
56	72.684402631991\\
57	72.684402631991\\
59	72.684402631991\\
65	72.684402631991\\
66	72.684402631991\\
70	72.684402631991\\
71	72.684402631991\\
73	72.684402631991\\
76	72.684402631991\\
80	72.684402631991\\
83	72.684402631991\\
86	72.684402631991\\
88	72.684402631991\\
89	72.684402631991\\
92	72.684402631991\\
93	72.684402631991\\
95	72.684402631991\\
101	72.684402631991\\
102	72.684402631991\\
103	72.684402631991\\
105	72.684402631991\\
113	72.684402631991\\
114	72.684402631991\\
117	72.684402631991\\
117	88.6881958660231\\
117	117.00259928008\\
119	117.688787745457\\
121	124.337162819038\\
123	124.337162819038\\
127	124.337162819038\\
128	124.337162819038\\
129	124.337162819038\\
130	124.388965388095\\
130	128.655048123862\\
135	191.984032594613\\
137	194.35120933063\\
138	196.064234536499\\
143	196.064234536499\\
148	196.064234536499\\
151	196.064234536499\\
153	196.301917899436\\
155	197.02156545024\\
158	197.02156545024\\
164	197.02156545024\\
167	197.02156545024\\
173	197.02156545024\\
175	197.02156545024\\
178	197.02156545024\\
182	197.02156545024\\
185	197.02156545024\\
186	197.02156545024\\
190	197.02156545024\\
197	197.02156545024\\
198	197.02156545024\\
199	197.02156545024\\
201	197.02156545024\\
204	197.02156545024\\
209	197.02156545024\\
210	197.02156545024\\
216	197.02156545024\\
224	197.02156545024\\
227	197.02156545024\\
229	197.02156545024\\
231	197.02156545024\\
235	218.177627613459\\
248	231.765004058571\\
253	268.748637167702\\
261	268.748637167702\\
265	268.748637167702\\
269	320.401397354748\\
270	320.401397354748\\
272	320.401397354748\\
281	320.401397354748\\
284	320.401397354748\\
303	333.879019822061\\
319	347.195243561687\\
339	385.889324477911\\
353	393.085799985951\\
};
\end{axis}

\end{tikzpicture}%
}
	\vspace{-0.3cm}
	\hspace{-0.3cm}\subfloat[S-IBP]{
%
%
\definecolor{mycolor1}{rgb}{0.00000,0.44700,0.74100}%
\begin{tikzpicture}

\begin{axis}[%
width=0.37\columnwidth,
height=0.22\columnwidth,
scale only axis,
xlabel style={at={(axis description cs:0.5,-0.03)},anchor=north},
every x tick label/.append style={font=\footnotesize},
xmin=0,
xmax=400,
xtick={  0, 400},
xlabel style={font=\footnotesize\color{white!15!black}},
xlabel={Empirical},
every y tick label/.append style={font=\footnotesize},
ymin=0,
ymax=400,
ytick={400},
ylabel style={font=\footnotesize\color{white!15!black}},
ylabel style={at={(axis description cs:-0.03,.5)},anchor=south},
ylabel={Inferred\tiny},
axis background/.style={fill=white},
axis x line*=bottom,
axis y line*=left
]
\addplot [color=red, dashdotted, forget plot]
  table[row sep=crcr]{%
3	15.1370116156597\\
361	357.185868311796\\
};
\addplot [color=red, forget plot]
  table[row sep=crcr]{%
54	63.8646420388523\\
186	189.98321489888\\
};
\addplot [color=mycolor1, draw=none, mark=+, mark options={solid, blue}, forget plot]
  table[row sep=crcr]{%
3	6.84477807618219\\
4	8.70928777258308\\
11	17.8314111561882\\
11	18.1309162855702\\
15	18.1589694347115\\
15	18.9160511607845\\
16	19.0286449615799\\
16	19.0515913206933\\
17	19.0974563303989\\
17	22.673665236907\\
19	27.8974254749681\\
20	29.3834125307525\\
23	29.7385122570753\\
27	29.8789804519486\\
28	33.1185870977863\\
29	33.7617084757082\\
29	40.6238173441938\\
33	41.5553970357512\\
34	41.5999011257017\\
35	48.1352241334208\\
36	50.1936004714437\\
37	54.7003976749995\\
41	54.7211393654609\\
42	55.5992218346949\\
43	57.6398999107927\\
44	57.9745232343294\\
44	58.5547186741426\\
45	61.7407206400314\\
50	63.3385310958712\\
52	63.4456121590392\\
54	63.8646420388523\\
56	66.0569952688131\\
57	66.5986374291651\\
59	70.3812469048601\\
65	77.8941661820882\\
66	78.0732526591415\\
66	78.6364188988496\\
70	80.485545001566\\
71	83.2357151634232\\
71	86.2764163635616\\
73	86.4518590024669\\
76	88.4527602124581\\
80	90.8418521089189\\
83	92.7009987534724\\
86	96.7190413453111\\
88	97.5026492162045\\
89	97.8879625460617\\
92	100.014531803982\\
93	100.526389544184\\
93	101.498642977784\\
93	103.082902798966\\
95	104.564732258984\\
95	105.176874603477\\
101	109.76507467358\\
102	110.992942091602\\
103	112.968337388165\\
103	118.607170393712\\
105	119.122310607641\\
105	119.712390415888\\
105	124.148749902766\\
113	125.707916787698\\
114	125.946095221286\\
117	126.410871050748\\
117	127.011641082517\\
117	127.188982503862\\
117	128.777627550138\\
119	131.903999643942\\
121	137.343877646999\\
121	137.649940530104\\
123	139.379758016333\\
127	141.042516579551\\
128	142.457658354491\\
129	142.608778013741\\
130	142.697729755276\\
130	147.614153895634\\
135	148.085710366563\\
137	149.38396983798\\
138	150.328034589211\\
143	155.033938886389\\
148	156.11219190755\\
151	156.574621515029\\
153	164.611102362315\\
155	165.560654728877\\
158	168.580751458739\\
164	168.686728434604\\
164	169.68714002766\\
167	173.081544009993\\
173	177.645747611982\\
175	178.757048153924\\
178	183.860749027479\\
182	184.16711664127\\
185	187.635476660943\\
186	189.98321489888\\
186	195.780219574746\\
190	198.358466149609\\
197	203.125579318546\\
198	209.406984521034\\
198	209.536149561242\\
199	209.916048502085\\
201	210.744069992941\\
204	214.997975814453\\
209	221.032979365813\\
210	222.55100278093\\
216	225.484930636862\\
224	227.93564066735\\
224	232.92204842131\\
227	233.740577618265\\
229	234.994364507958\\
231	235.467971471864\\
235	237.304839440545\\
248	239.065564731679\\
253	240.179883820638\\
261	243.651406882347\\
265	243.804338204337\\
269	244.77438792556\\
270	245.542296261688\\
272	249.777107865879\\
281	251.107996310073\\
284	252.915555245995\\
303	257.631649308283\\
319	260.934218594501\\
339	271.980255713565\\
353	273.064326798221\\
361	279.548552594869\\
};
\end{axis}

\end{tikzpicture}%
}\subfloat[S3R-IBP]{
%
%
\definecolor{mycolor1}{rgb}{0.00000,0.44700,0.74100}%
\begin{tikzpicture}

\begin{axis}[%
width=0.37\columnwidth,
height=0.22\columnwidth,
scale only axis,
xlabel style={at={(axis description cs:0.5,-0.03)},anchor=north},
every x tick label/.append style={font=\footnotesize},
xmin=0,
xmax=400,
xtick={  0, 400},
xlabel style={font=\footnotesize\color{white!15!black}},
xlabel={Empirical},
every y tick label/.append style={font=\footnotesize},
ymin=0,
ymax=400,
ytick={400},
ylabel style={font=\footnotesize\color{white!15!black}},
ylabel style={at={(axis description cs:-0.03,.5)},anchor=south},
ylabel={Inferred\tiny},
axis background/.style={fill=white},
axis x line*=bottom,
axis y line*=left
]
\addplot [color=red, dashdotted, forget plot]
  table[row sep=crcr]{%
3	0.970157480216983\\
361	361.167686618372\\
};
\addplot [color=red, forget plot]
  table[row sep=crcr]{%
54	52.2832133071609\\
186	185.093475447486\\
};
\addplot [color=mycolor1, draw=none, mark=+, mark options={solid, blue}, forget plot]
  table[row sep=crcr]{%
3	52.0723974061869\\
4	52.0723974061869\\
11	52.0723974061869\\
15	52.0723974061869\\
16	52.0723974061869\\
17	52.0723974061869\\
19	52.0723974061869\\
20	52.0723974061869\\
23	52.0723974061869\\
27	52.0723974061869\\
28	52.0723974061869\\
29	52.0723974061869\\
33	52.0723974061869\\
34	52.0723974061869\\
35	52.0723974061869\\
36	52.0723974061869\\
37	52.0723974061869\\
41	52.0723974061869\\
42	52.0723974061869\\
43	52.0723974061869\\
44	52.0723974061869\\
45	52.0723974061869\\
50	52.0723974061869\\
52	52.0723974061869\\
54	52.2832133071609\\
56	60.8025828155809\\
57	65.8701778042639\\
59	65.9755748123318\\
65	69.7809330879992\\
66	73.3648034045565\\
70	73.3648034045565\\
71	75.1808572172242\\
71	75.8345906022086\\
73	76.1629851781329\\
76	77.0501518348082\\
80	81.5901366123301\\
83	81.5901366123301\\
86	81.5901366123301\\
88	81.5901366123301\\
89	81.742917980202\\
92	82.6936548362477\\
93	83.6982956220697\\
93	86.0977183699662\\
93	88.4680565087791\\
95	89.3903249483665\\
95	90.8326420641424\\
101	99.8129623204889\\
102	99.8953075816906\\
103	103.124924368555\\
103	103.369344662972\\
105	103.474398574758\\
105	103.504849482503\\
105	104.579214915688\\
113	104.854545325801\\
114	107.383222886944\\
117	118.160679397895\\
117	121.277445599851\\
117	123.057143077024\\
117	124.361096427666\\
119	125.951699021742\\
121	126.460058984623\\
121	127.285471137742\\
123	130.533154829861\\
127	132.030401692621\\
128	134.004530450439\\
129	135.406872416742\\
130	136.377165359931\\
130	136.51881184905\\
135	137.573201086101\\
137	137.786740256799\\
138	141.115144262904\\
143	142.369073624717\\
148	142.979373154779\\
151	144.560693078159\\
153	146.214508283028\\
155	149.43273633945\\
158	154.424454036967\\
164	159.133600091244\\
164	161.529971442914\\
167	165.425188249431\\
173	166.578537228854\\
175	173.527265539804\\
178	176.870492716302\\
182	180.509612163514\\
182	184.587018619923\\
185	184.769011246584\\
186	185.093475447486\\
186	185.273094829734\\
190	194.510160443128\\
197	194.948563166938\\
198	212.278760878218\\
198	212.529638405765\\
199	213.016300754115\\
201	214.094795003531\\
204	214.927332330543\\
209	215.311091100835\\
210	216.780454353943\\
216	221.322326482955\\
224	222.118333051501\\
224	222.936131270423\\
227	225.890780752548\\
229	225.903409743266\\
231	229.813580085663\\
235	233.171228710178\\
248	235.524700334277\\
253	237.164811616296\\
261	238.184526019703\\
265	250.052017548713\\
269	257.25373599262\\
270	258.208599858168\\
272	263.169498942287\\
281	266.1883938194\\
284	270.179944878473\\
303	280.367974282485\\
319	300.879412603155\\
339	301.065585324831\\
353	313.670221562628\\
361	317.867031781854\\
};
\end{axis}

\end{tikzpicture}%
}
	\caption{\textbf{Capturing sparsity structure.} S3R-IBP gives the best fit for the distribution of number of non-zero values per row in $\X$.}
	\vspace{-0.6cm}
	\label{fig:diversity}
\end{figure}

\vspace{-0.1cm}
\paragraph*{Sparsity structure}
Figure~\ref{fig:diversity} evaluates the ability of S3R-IBP to fit the input distribution of the number of non-zero values per-row in~$\X$, versus IBP, S-IBP, and a simple binomial model from the economic literature~\cite{hausmann_network_2011}. 
We measure the ``proximity'' of the empirical and predicted distribution via qq-plots.
S-IBP underfits the distribution for higher values, e.g., it predicts a lower number of countries with high number of exports, in contrast to the S3R-IBP model.

%

\begin{table*}
\scriptsize
\vspace{-0.6cm}
\begin{minipage}{0.42\textwidth}
\vspace{0.25cm}
  \centering
    \hspace{-0.4cm}%
    \begin{tabular}{cC{6.9cm}}
    
    \toprule
{ \bf Id } & { \bf Products with highest weights } \\
\hline \\
{\color{blue} F1}   & {\color{blue} misc. animal oils (0.78), bovine entails (0.72), bovine meat (0.68), milk (0.63), equine (0.62), butter (0.58) }\\
F2   & synthetic woven, synth. yarn, woven $<$ 85\% synth.\\
F3   & parts metalworking, tool parts, polishing stones\\
F4   & Aldehyde--Ketone, glycosides--vaccines, medicaments\\
F5   & synthetic rubber, acrylic polymers, silicones\\
F6   & measuring instruments, math inst., electrical inst.\\
F7   & vehicles parts, cars, iron wire\\
F8   & improved wood, mineral wool, heating equipment\\
F9   & elect. machinery, vehicles stereos, data processing eq.\\
F10   & baked goods, metal containers, misc. edibles\\
F11  & misc. articles of iron, carpentry wood, wood articles\\
F12   & vegetables, fruit--vegetable juices, misc. fruit\\
F13   & misc. pumps, ash--residues, chemical wood pulp\\
F14   & synth. undergarments, feminine outerwear, men's shirts \\
F15   & misc. rotating, electric plant parts, control inst. of gas\\
                                 \toprule
\end{tabular}
  \caption{\textbf{Features learned by S3R-IBP.} Products with higher weights are reported.}\label{tab:all_topics}
\end{minipage}
\hfill
\hfill
\begin{minipage}{0.21\textwidth}
\vspace{0.2cm}
\begin{tabular}{c}
 \toprule
{\textbf{IBP}} \\
\hline \\
confectionary sugar (0.45) \\
plastic containers (0.43)  \\
baked goods (0.41)    \\
tissue paper (0.40)      \\
metal containers (0.39)\\
soaps (0.39)\\
\toprule
\end{tabular}

\vspace{0.15cm}
\begin{tabular}{c}
 \toprule
{\textbf{S-IBP}} \\
\hline \\
bovine (0.53) \\
improved wood (0.51)  \\
misc. vegetable oils (0.50)  \\
butter (0.50)  \\
rape seeds (0.47) \\
misc. wheat (0.45) \\
\toprule
\end{tabular}
  \caption{\textbf{Competitors.} Example matched to F1.}\label{tab:example_topic}
\end{minipage}
\begin{minipage}{0.28\textwidth}
\scriptsize
  \centering
 \subfloat[M-F0]{%
    \hspace{0.2cm}%
    \begin{tabular}{cc}
    
    \toprule
{ \bf Id } & { \bf Weight } \\
\hline \\
\color{red}     F14  &	0.37 \\
\color{red}     F12	  & 0.32 \\
\color{green}     F10 &	0.17 \\
\color{red}     F2 &	0.16  \\
\color{red}     F1 &	0.14  \\
\color{red}     F9 &	0.13  \\
     F13 & 	0.05 \\
     F6  &	0.04 \\
     F5  &	0.04 \\
     F4  &	0.04 \\ 
     F15  &	0.04 \\ 
     F7  &	0.03 \\ 
     F8  &	0.03 \\ 
     F11  &	0.02 \\
     F3  &	0.02 \\ 
                                 \toprule
\end{tabular}
  }
  \subfloat[M-F1]{%
    \hspace{.2cm}%
    \begin{tabular}{cc}
    \toprule
{ \bf Id} & { \bf Weight} \\
\hline \\
F8 &	0.69 \\
F11&0.68 \\
F15&	0.60 \\
\color{green} F10& 0.59 \\
F7&	0.52 \\
F6&	0.34 \\
F13&	0.32 \\
F4	&0.31 \\
F3&	0.31\\
F5&	0.14\\
\color{red} F1&	0.05\\
\color{red} F9&	0.02\\
\color{red} F2&	0.01\\
\color{red} F14&	0.00\\
\color{red} F12&	0.00\\
                                 \toprule
\end{tabular}
  }
  \vspace{-0.3cm}
  \caption{\textbf{Meta-features.} A sharp division of the world arises.}\label{tab:deepIBP}
\end{minipage}
\vspace{-0.5cm}
\end{table*}

\subsection{Qualitative evaluation}

\paragraph*{Interpretability}
Table~\ref{tab:all_topics} lists all the averaged latent features learned by S3R-IBP. Table~\ref{tab:example_topic} shows feature example F1 learned by IBP and S-IBP. Features are matched across the 10-folds using the Jaccard index similarity. S3R-IBP is able to give much shorter and concise descriptions, as weights decrease at a faster pace and reach higher values at the top. Products in the IBP list are heterogeneous. The S-IBP list includes items from a mixture of farming and technological elements, whereas the S3R-IBP list is more homogeneous.

\paragraph*{Features correlation}
To analyze the existing correlation between latent features, we apply our S3R-IBP on $\Z$ as input data. Such a deep structure, i.e., using a two-layer IBP, has already been explored in~\cite{doshi-velez_correlated_2009}. 
S3R-IBP infers two meta-features, M-F0 and M-F1, which assign different weights to each latent feature from the first layer. Countries with an active M-F1 are those that have more active features in the first layer and larger GDP. M-F1 can be interpreted as the meta-feature that distinguishes between developed countries and developing ones, resulting in a sharp division of the world in terms of capabilities. 

M-F0 and M-F1 divide the original features into three disjoint sets.
 The first set contains features whose weight is either zero or insignificant in M-F1 (highlighted in red). These define countries with least capabilities, dealing with less complex products like farming or textile (see Table~\ref{tab:all_topics}).
%
%
 The second set is composed by F10 (in green), which has a high value in both meta-features. This feature is present in both developing and developed countries, although developed countries do trade them more efficiently than developing ones (higher weights). 
%
 The last set includes features whose weights in M-F0 are negligible compared to their weights in M-F1 (in black). These features contain products like chemicals and complex machinery, which are mostly traded by developed countries.
%

 Such sharp division among features suggests the existence of a ``poverty'' or ``quiescence trap'' in the spirit of~\cite{hausmann_network_2011}, a trap of development stasis in which some countries get stuck due to the inability to ``acquire'' capabilities associated with the production of more complex products. 

\section{Conclusion}

This paper presents the S3R-IBP model, a non-negative  and sparse infinite matrix factorization for data exploration of high-dimensional count data.
The model is able to capture complex sparsity structure, and delivers compact, easy to interpret features.
 We illustrate the usefulness of this model to explain trade data, by interpreting latent features as country capabilities which are required for producing products.
%
%
The presented approach is general for any other count-data scenario where broad, flexible assumptions are needed.
 As future work, we plan to introduce a Markovian dependency in the latent space, allowing for dynamic feature activation over time.


\end{document}